\documentclass{article}
\usepackage{spconf,amsmath,graphicx}
\usepackage{color}

\usepackage{amsfonts}
\usepackage{pifont}
\usepackage{bm}
\usepackage{comment}
\usepackage{xcolor}
\usepackage{url}
\usepackage[capitalize]{cleveref}
\usepackage{multirow}
\usepackage{array,booktabs}
\usepackage{enumerate}
\usepackage[shortlabels]{enumitem}

\renewcommand{\vec}[1]{\bm{#1}}

\newcommand{\mat}[1]{\bm{\mathrm{#1}}}

\newcommand{\nm}[1]{{\mathtt{#1}}}

\newcommand{\nnet}[1]{\mathcal{#1}}
\newcommand{\ember}{\nnet{M}}
\newcommand{\vocab}{V}

\newcommand{\thinnerspace}{\hspace{0.1em}}

\crefformat{section}{\S#2#1#3} 
\crefformat{subsection}{\S#2#1#3}
\crefformat{subsubsection}{\S#2#1#3}

\title{MULTIPLE REPRESENTATION TRANSFER FROM LARGE LANGUAGE MODELS \\TO END-TO-END ASR SYSTEMS}
\name{Takuma Udagawa \enspace Masayuki Suzuki \enspace Gakuto Kurata \enspace Masayasu Muraoka \enspace George Saon}
\address{IBM Research AI}
\begin{document}
%
\maketitle
\begin{abstract}
Transferring the knowledge of large language models (LLMs) is a promising technique to incorporate linguistic knowledge into end-to-end automatic speech recognition (ASR) systems.
However, existing works only transfer a single representation of LLM (e.g. the last layer of pretrained BERT), while the representation of a text is inherently non-unique and can be obtained variously from different layers, contexts and models.
In this work, we explore a wide range of techniques to obtain and transfer multiple representations of LLMs into a transducer-based ASR system.
While being conceptually simple, we show that transferring multiple representations of LLMs can be an effective alternative to transferring only a single LLM representation.
\end{abstract}
\section{Introduction}
\label{sec:intro}

\begin{figure*}[t]
\centering
\includegraphics[width=0.9\textwidth]{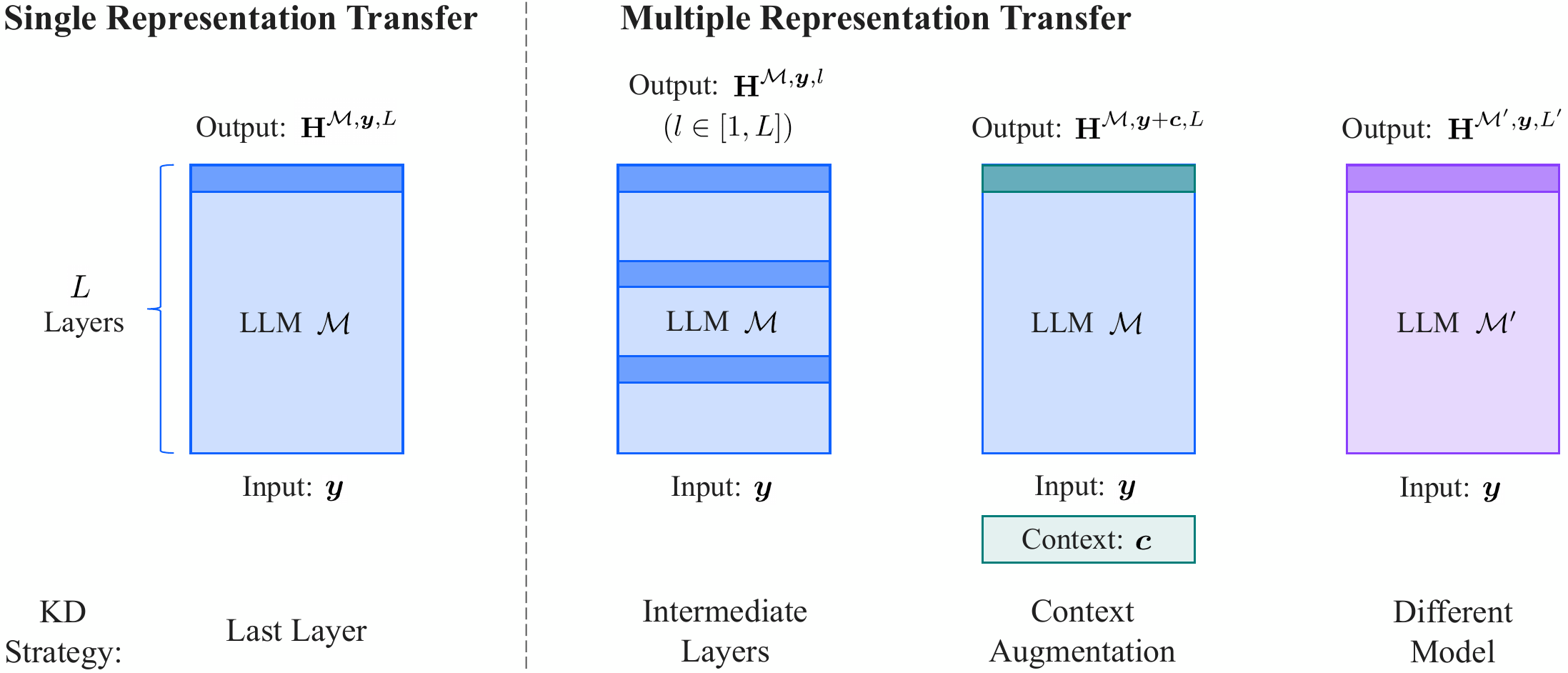}
\caption{Illustration of the single (left) and multiple (right) representation transfer. In multiple representation transfer, we obtain and transfer multiple representations of LLM from intermediate layers, context augmentation and different model (\cref{ssec:multiple_representation_transfer}).
}
\label{fig:multiple_transfer}
\end{figure*}

Large language models (LLMs) pretrained on massive text corpora have become a dominant paradigm in NLP \cite{devlin-etal-2019-bert,conneau-etal-2020-unsupervised,OpenAI2023GPT4TR}.
Based on a stack of Transformer layers \cite{vaswani2017attention}, LLMs progressively learn high-level representations of the input text which can be useful for solving various linguistic tasks \cite{wang2019glue,hu2020xtreme}.
When the model size and training data are scaled up sufficiently, LLMs are known to exhibit further emergent abilities such as few-shot learning and instruction following \cite{NEURIPS2020_1457c0d6,ouyang2022training}.

In the context of automatic speech recognition (ASR), several works have explored integrating the LLMs into end-to-end ASR systems to enhance its linguistic proficiency and improve transcription accuracy \cite{salazar-etal-2020-masked,huang2021speech,yi2021efficiently,higuchi2023bectra,rubenstein2023audiopalm}.
Amongst other approaches, knowledge distillation \cite{Hinton2015DistillingTK} is particularly appealing since it does not require any extra inference cost or modification to the ASR system architecture.
Existing works have proposed either transferring the predictive distribution \cite{futami2020distilling} or layer representation \cite{bai2021fast,kubo2022knowledge,lu2022context,deng2022improving,han2023knowledge} of the LLM (teacher) to the ASR system (student).
From our preliminary study, we found the latter approach to be more promising and hence focus on this setting in our experiments.

In the layer representation transfer, the ASR system is trained with an auxiliary objective of predicting the linguistic features present in the LLM representation, e.g. the last layer of pretrained BERT \cite{devlin-etal-2019-bert}. 
However, such representation can be obtained in various ways, capturing different and complementary aspects of the input text.
Firstly, intermediate layers of LLM empirically capture more local/syntactic features \cite{tenney-etal-2019-bert} and can facilitate distillation by bridging the capacity gap between the teacher and student \cite{sun-etal-2019-patient}.
Secondly, we can provide additional context (e.g. past and future utterances) as input to make the representations more informative \cite{liu-etal-2022-rethinking-task}.
Finally, we can naturally use different models, e.g. trained on different datasets or based on different architectures, to obtain various representations of the input text.
Figure 1 summarizes our approaches of obtaining multiple layer representations.

In this work, we study whether transferring multiple representations of LLM can improve ASR performance.
Specifically, we experiment on the commonly used Switchboard corpus with a Conformer-Transducer (Cfm-T) \cite{Gulati2020Conformer} as the ASR baseline and transfer the knowledge of BERT-family models\footnote{While BERT models are smaller in size ($\leq$340M parameters) compared to the latest LLMs ($>$100B parameters) \cite{NEURIPS2020_1457c0d6,OpenAI2023GPT4TR}, we expect them to be better teacher with smaller capacity gap for our ASR baseline (71M parameters).} by extending Kubo et al.'s method \cite{kubo2022knowledge}.
We conduct extensive evaluation on the 300 hours corpus and extend it to the larger 2000 hours corpus based on the best configuration.
Overall, we show that multiple representation transfer can be an effective alternative to single representation transfer, especially on the 300 hours setting with limited amount of training data.

\section{Related Work}
\label{sec:related_work}

Current end-to-end ASR systems \cite{prabhavalkar2017comparison} require paired speech and text transcripts for training, limiting the usage of unpaired text corpora to acquire linguistic knowledge.
To complement this, external language models have been applied during first-pass decoding \cite{toshniwal2018comparison,mcdermott2019density} or second-pass rescoring of the N-best hypotheses \cite{salazar-etal-2020-masked,shin2019effective,Chiu2021InnovativeBR}, successfully improving ASR accuracy on the most competitive baselines \cite{udagawa2022effect}.
LLMs have also been applied as a postprocessing to correct grammatical or semantic errors in the predicted transcriptions \cite{zhao2021bart,min2023exploring}.

Another line of research aims to unify the LLM and the ASR system into a single architecture.
A majority of approaches focus on creating a shared embedding space for the text and speech modalities, so that the LLM module can consume the acoustic features to assist predictions \cite{huang2021speech,yi2021efficiently,rubenstein2023audiopalm,zheng-etal-2021-wav-bert}.
Alternatively, one can leverage the pretrained LLM without any finetuning to iteratively refine the hypotheses \cite{higuchi2023bectra,higuchi-etal-2022-bert}.

In contrast to the above approaches, knowledge distillation does not incur any additional inference cost nor modification to the system architecture.
However, prior works focus on transferring only a single LLM representation, i.e. the last layer \cite{bai2021fast,kubo2022knowledge,deng2022improving,han2023knowledge} or the average of all layers \cite{lu2022context,liu2020sequence}. 
Therefore, the benefit of transferring multiple representations as individual features has not been explored.
In this study, we conduct a comprehensive analysis of transferring multiple LLM representations into a transducer ASR baseline.

\section{Methods}
\label{sec:methods}

Let $\mat{X} = (\vec{x}_1,\vec{x}_2,...)$ denote the input acoustic features. 
In a transducer-based ASR system \cite{graves2012sequence}, the input $\mat{X}$ is encoded by a transcription network to produce the acoustic features $\vec{\phi} = (\vec{\phi}_1,...,\vec{\phi}_T)$, where each $\vec{\phi}_t \in \mathbb{R}^{D^{\nm{Trs}}}$.
Alongside, a hypothesis $\vec{y} = (y_1,...,y_N)$ can be encoded by a prediction network, where each $y_i$ is a token from a vocabulary set $V$ which we assume to be shared by the LLM.\footnote{Specifically, we use BERT's wordpiece vocabulary in this study.}
The prediction network outputs the text features $\vec{\psi} = (\vec{\psi}_1,..., \vec{\psi}_N)$, where each $\vec{\psi}_i \in \mathbb{R}^{D^{\nm{Prd}}}$ is conditioned only on the prefix $\vec{y}_{1:i-1}$: this is important for the next token $y_i$ to be predicted based on $\vec{\psi}_i$ (i.e. the prefix $\vec{y}_{1:i-1}$) and the hypothesis to be generated in an auto-regressive manner.

Based on $\vec{\phi}$ (and whilst encoding $\vec{\psi}$), a joint network computes the probability distribution over the alignment variables $\vec{z} = (z_1,...,z_{T + N})$, where $z_i \in \vocab \cup \left\{ \phi \right\}$ and $\phi$ denotes a blank symbol.
By marginalizing out the alignment variable, we can compute the probability distribution over the hypotheses $p(\vec{y} \mid \mat{X})$.
For further details on model training and inference, we refer to \cite{graves2012sequence,saon2021advancing}.
During ASR training, the network parameters are updated to maximize the probability of gold transcriptions based on the ASR training loss $\mathcal{L}_{\textrm{ASR}}$.

\subsection{Single Representation Transfer}
\label{ssec:single_representation_transfer}

Next, we describe the method of transferring a single representation of LLM into the transducer baseline, following \cite{kubo2022knowledge}.
Let $\ember$ denote the LLM with $L$ Transformer layers.
Given the input text $\vec{y}$, let $\mat{H}^{\ember,\vec{y},L}\hspace{-0.2em}= (\vec{h}^{\ember,\vec{y},L}_1\hspace{-0.2em},...,\vec{h}^{\ember,\vec{y},L}_N)$ denote the last ($L$-th) layer representation (or hidden state) of $\ember$, where $\vec{h}^{\ember,\vec{y},L}_i\hspace{-0.2em} \in \mathbb{R}^{D^{\nm{Hid}}}$.
For brevity, we omit the superscripts unless necessary, i.e. use the notations $\mat{H}$ and $\vec{h}_i$.

Based on the obtained layer representation, the acoustic and text features $\vec{\phi}_t$ and $\vec{\psi}_i$ are trained to predict the linguistic features present in $\vec{h}_i$.
Specifically, we use a regression network $\nnet{R}: \mathbb{R}^{D^{\nm{Trs}}+D^{\nm{Prd}}} \to \mathbb{R}^{D^{\nm{Hid}}}$ to predict $\vec{h}_i$ from $\vec{\phi}_t$ and $\vec{\psi}_i$.
The precise loss function can be defined as follows:
\begin{equation}
\begin{aligned}
    \mathcal{L}_{\textrm{KD}} &= \sum_{i \in [1,N]}
    \mathbb{E}_{\thinnerspace q_i(t \mid \mat{X}, \vec{y})} \thinnerspace \bigr[ \thinnerspace
    d \thinnerspace(
    \nnet{R} ( \vec{\phi}_t, \vec{\psi}_{i}) ,
    \vec{h}_i ) \thinnerspace \bigr] \thinnerspace\\
    &\simeq \sum_{i \in [1,N]}
    d \thinnerspace \bigl(
    \mathcal{R} (\thinnerspace 
    \mathbb{E}_{\thinnerspace q_i(t \mid \mat{X}, \vec{y})}  \thinnerspace [ \thinnerspace
    \vec{\phi}_t \thinnerspace]\thinnerspace, \vec{\psi}_{i}) ,
    \vec{h}_i \thinnerspace \bigr)
\end{aligned}
\label{eq:kd_loss}
\end{equation}
\noindent
Here, $d(.)$ is the distance function such as the $\mathrm{L}^p$ distance. The alignment probability $q_i(t \mid \mat{X}, \vec{y})$ is the probability of comsuming the $t$-th acoustic feature for predicting the token $\vec{y}_i$, which can be computed in the process of marginalizing $\vec{z}$.
The approximation is accurate when L$^2$ distance and a linear network are used for $d(.)$ and $\mathcal{R}$, respectively \cite{kubo2022knowledge}.

Overall, the loss function including both ASR training and knowledge distillation from LLM is defined as $\mathcal{L}_{\textrm{ASR}} + \lambda \cdot \mathcal{L}_{\textrm{KD}}$, where the weight hyperparameter $\lambda$ is set empirically.

\subsection{Multiple Representation Transfer}
\label{ssec:multiple_representation_transfer}

Extending the previous approach, we first obtain multiple representations of LLM based on the following procedure:

\noindent
\textbf{Intermediate Layers}\quad
In the distillation of LLMs, transferring not only the last layer but also the intermediate layers has been shown to improve performance \cite{sun-etal-2019-patient,udagawa-etal-2023-comparative}.
Here, the concept of layer mapping strategy, i.e. which set of layers to transfer, plays a key role.
In this paper, we conduct a comprehensive evaluation comparing the following strategies:
\begin{itemize}[topsep=0pt, itemsep=0pt, leftmargin=.2in, parsep=0pt]
\item \textit{Last}: We select the last $K$ layers of LLM, which empirically capture higher-level semantics of the text \cite{tenney-etal-2019-bert}.
\item \textit{First}: We select the first $K$ layers of LLM, which empirically capture more local/syntactic features \cite{tenney-etal-2019-bert}.
\item \textit{Uniform}: We select every $k$ layers totaling $K$ layers (i.e. $k = \lceil L/K \rceil$) to cover both higher and lower layers \cite{jiao-etal-2020-tinybert}.
\item \textit{Random}: We randomly select $K$ layers in each epoch, which can potentially work as a regularizer \cite{haidar-etal-2022-rail}.
\end{itemize}
In our experiments, we use $K \in \{1,2,3,4\}$ layers for each strategy. We also experiment with the mean pooling of all layers \cite{lu2022context,liu2020sequence} as an existing baseline.

\noindent
\textbf{Context Augmentation}\quad
We can also provide additional context $\vec{c}$ as input to make the representation more informative, e.g. by concatenating relevant sentences retrieved from an external corpus \cite{liu-etal-2022-rethinking-task}.
In this study, we take a simpler approach and concatenate the past and future utterances before and after $\vec{y}$, respectively.
We use a fixed context size of 60 tokens and apply random masking of the context tokens \textit{on the fly} to encourage representation diversity.

\noindent
\textbf{Different Model}\quad
Finally, we can naturally use a different model $\ember^\prime$, trained on a different dataset or based on a different architecture, to obtain the representation $\mat{H}^{\ember^\prime,\vec{y},L^\prime}$\hspace{-0.2em}.
In this study, we use BERT-Base as the primary model $\ember$ and combine it with a model $\ember^\prime$ from the following set:
\begin{itemize}[topsep=0pt, itemsep=0pt, leftmargin=.2in, parsep=0pt]
\item \textit{BERT-Finetuned}: This is a 12 layer BERT-Base model adapted to the speech domain by finetuning on the transcribed text of 2000 hours of Switchboard corpus.
\item \textit{DistilBERT}: This is a smaller, 6 layer BERT model distilled on English Wikipedia and BookCorpus \cite{sanh2019distilbert}.
\item \textit{BERT-Large}: This is a larger, 24 layer BERT model pretrained on English Wikipedia and BookCorpus \cite{devlin-etal-2019-bert}.
\end{itemize}
Note that BERT-Base and BERT-Finetuned are only different in the training data (i.e. based on the same architecture), while BERT-Base, DistilBERT and BERT-Lage are only different in the model architecture (i.e. trained on the same data).

\medskip
Based on the above procedure, we can obtain multiple representations of LLM denoted as $\vec{h}^{\textrm{multi}}_i = (\vec{h}^1_i,\vec{h}^2_i,...)$ with a total dimension size of $D^{\nm{HidMulti}}$.\footnote{We concatenate each representation $\vec{h}^j_i$ as individual features instead of aggregating them into a single representation (e.g. based on mean pooling).}
This can be transferred to the transducer ASR system in the same way as single representation transfer, i.e. by replacing $\vec{h}_i$ with $\vec{h}^{\textrm{multi}}_i$ and $\nnet{R}$ with $\nnet{R}^{\textrm{multi}}: \mathbb{R}^{D^{\nm{Trs}}+D^{\nm{Prd}}} \to \mathbb{R}^{D^{\nm{HidMulti}}}$ in eq. (\ref{eq:kd_loss}).

\section{Experiments}
\label{sec:experiments}

\subsection{Experimental Setup}
\label{ssec:experimental_setup}

In our experiments, we assess the effect of LLM knowledge distillation on a Conformer Transducer (Cfm-T) baseline.
For the training of our Cfm-T models, we use the widely adopted 300 or 2000 hours of the Switchboard corpus, a dyadic English telephone conversation dataset.

For acoustic features, we use 40-dimensional log-Mel filterbank energies, their delta, and double-delta coefficients, stacking consecutive frames and skipping every other frame to produce 240-dimensional features.
The transcription network consists of 6 Conformer blocks, which encodes the input into 640-dimensional accoustic features.
For text features, we tokenize the input text with BERT tokenizer and encode it with a unidirectional LSTM into 768-dimensional features.
Based on these features, the joint network is applied following the multiplicative integration method \cite{saon2021advancing}.

Our models are trained for a total of 2 iterations, each containing 20 epochs.
In the 1$^{\textrm{st}}$ iteration, we only use the ASR training loss $\mathcal{L}_{\textrm{ASR}}$.
In the 2$^{\textrm{nd}}$ iteration, we use the combined loss $\mathcal{L}_{\textrm{ASR}} + \lambda \cdot \mathcal{L}_{\textrm{KD}}$ with $\lambda=0.01$.
We fix the alignment probability $q_i(t \mid \mat{X}, \vec{y})$ in eq. (\ref{eq:kd_loss}) based on the 1$^{\textrm{st}}$ iteration model to stabilize distillation.
Following \cite{kubo2022knowledge}, we use $\mathrm{L}^1$ as the distance function $d(.)$ and a linear projection as $\mathcal{R}$.

First, we conduct extensive experiments on the 300 hours corpus, running the 2$^{\textrm{nd}}$ iteration with various configurations and 3 times with different random seeds.
Then, we extend it to the larger 2000 hours corpus based on the best configuration. We evaluate the models on the commonly used Hub5 2000 Switchboard (SWB) and CallHome (CH) test sets and report the Word Error Rate (WER) as the evaluation metric.\footnote{For the 300 hours experiments, we report the mean and standard deviation of the WERs from the 2$^{\textrm{nd}}$ iteration run with 3 different random seeds.}

\subsection{Results}
\label{ssec:results}

\begin{table}[t]
\centering
\setlength{\aboverulesep}{1pt}
\setlength{\belowrulesep}{1pt}
\setlength\tabcolsep{8pt}
\begin{tabular}{lccc}
\toprule[\heavyrulewidth]
\multicolumn{2}{c}{KD Strategy} & SWB & CH \\
\midrule
\midrule
\multicolumn{2}{l}{No KD (1$^{\textrm{st}}$ itr)} & 21.0 & 12.2 \\
\multicolumn{2}{l}{No KD (2$^{\textrm{nd}}$ itr)} & 19.4{\scriptsize $\pm$0.2} & 11.0{\scriptsize $\pm$0.2} \\
\midrule
\multirow{4}{*}{Last} & 1L & 18.5{\scriptsize $\pm$0.1} & 10.4{\scriptsize $\pm$0.1} \\
& 2L & 18.4{\scriptsize $\pm$0.0} & 10.3{\scriptsize $\pm$0.2} \\
& 3L & \textbf{18.2{\scriptsize $\pm$0.0}} & \textbf{10.2{\scriptsize $\pm$0.1}} \\
& 4L & 18.6{\scriptsize $\pm$0.1} & 10.3{\scriptsize $\pm$0.1} \\
\midrule
\multirow{4}{*}{Uniform} & 1L & 18.5{\scriptsize $\pm$0.1} & 10.4{\scriptsize $\pm$0.1} \\
& 2L & 18.6{\scriptsize $\pm$0.2} & \textbf{10.1{\scriptsize $\pm$0.1}} \\
& 3L & \textbf{18.2{\scriptsize $\pm$0.2}} & \textbf{10.1{\scriptsize $\pm$0.1}} \\
& 4L & 18.3{\scriptsize $\pm$0.1} & \textbf{10.1{\scriptsize $\pm$0.1}} \\
\midrule
\multirow{4}{*}{First} & 1L & 18.7{\scriptsize $\pm$0.1} & \textbf{10.1{\scriptsize $\pm$0.1}} \\
& 2L & \textbf{18.4{\scriptsize $\pm$0.1}} & \textbf{10.1{\scriptsize $\pm$0.1}} \\
& 3L & \textbf{18.4{\scriptsize $\pm$0.0}} & 10.2{\scriptsize $\pm$0.1} \\
& 4L & \textbf{18.4{\scriptsize $\pm$0.0}} & 10.3{\scriptsize $\pm$0.1} \\
\midrule
\multirow{4}{*}{Random} & 1L & 18.5{\scriptsize $\pm$0.1} & \textbf{10.2{\scriptsize $\pm$0.0}} \\
& 2L & 18.4{\scriptsize $\pm$0.1} & \textbf{10.2{\scriptsize $\pm$0.0}} \\
& 3L & 18.4{\scriptsize $\pm$0.2} & \textbf{10.2{\scriptsize $\pm$0.2}} \\
& 4L & \textbf{18.3{\scriptsize $\pm$0.1}} & 10.3{\scriptsize $\pm$0.0} \\
\midrule
Mean Pool & 12L & 18.4{\scriptsize $\pm$0.1} & 10.1{\scriptsize $\pm$0.1} \\
\bottomrule[\heavyrulewidth]
\end{tabular}
\caption{
Results on Switchboard 300 hours training distilling various intermediate layers of pretrained BERT-Base.
}
\label{tab:swb_300_intermediate}
\end{table}

\begin{table}[t]
\centering
\setlength{\aboverulesep}{1pt}
\setlength{\belowrulesep}{1pt}
\setlength\tabcolsep{8pt}
\begin{tabular}{lcc}
\toprule[\heavyrulewidth]
KD Strategy & SWB & CH \\
\midrule
\midrule
Uniform 1L & 18.5{\scriptsize $\pm$0.1} & 10.4{\scriptsize $\pm$0.1} \\
\phantom{Unif}+ Context (0\% Mask) & 18.5{\scriptsize $\pm$0.3} & 10.3{\scriptsize $\pm$0.1} \\
\phantom{Unif}+ Context (10\% Mask) & \textbf{18.3{\scriptsize $\pm$0.2}} & \textbf{10.1{\scriptsize $\pm$0.1}} \\
\midrule
Uniform 2L & 18.6{\scriptsize $\pm$0.2} & 10.1{\scriptsize $\pm$0.1} \\
\phantom{Unif}+ Context (0\% Mask) & 18.2{\scriptsize $\pm$0.2} & 10.0{\scriptsize $\pm$0.1} \\
\phantom{Unif}+ Context (10\% Mask) & \textbf{18.1{\scriptsize $\pm$0.2}} & \textbf{9.8{\scriptsize $\pm$0.0}} \\
\bottomrule[\heavyrulewidth]
\end{tabular}
\caption{
Results on Switchboard 300 hours training combining the representation with additional context.
}
\label{tab:swb_300_context}
\end{table}

\begin{table}[t]
\centering
\setlength{\aboverulesep}{1pt}
\setlength{\belowrulesep}{1pt}
\setlength\tabcolsep{8pt}
\begin{tabular}{lcc}
\toprule[\heavyrulewidth]
KD Strategy & SWB & CH \\
\midrule
\midrule
Uniform 1L & 18.5{\scriptsize $\pm$0.1} & 10.4{\scriptsize $\pm$0.1} \\
\phantom{Unif}+ BERT-Finetuned & \textbf{18.4{\scriptsize $\pm$0.1}} & 10.3{\scriptsize $\pm$0.0} \\
\phantom{Unif}+ DistilBERT & \textbf{18.4{\scriptsize $\pm$0.0}} & \textbf{10.1{\scriptsize $\pm$0.1}} \\
\phantom{Unif}+ BERT-Large & 18.7{\scriptsize $\pm$0.0} & 10.3{\scriptsize $\pm$0.1} \\
\midrule
Uniform 2L & 18.6{\scriptsize $\pm$0.2} & 10.1{\scriptsize $\pm$0.1} \\
\phantom{Unif}+ BERT-Finetuned & \textbf{18.1{\scriptsize $\pm$0.0}} & 10.1{\scriptsize $\pm$0.0} \\
\phantom{Unif}+ DistilBERT & 18.2{\scriptsize $\pm$0.1} & \textbf{10.0{\scriptsize $\pm$0.2}} \\
\phantom{Unif}+ BERT-Large & 18.4{\scriptsize $\pm$0.1} & \textbf{10.0{\scriptsize $\pm$0.1}} \\
\bottomrule[\heavyrulewidth]
\end{tabular}
\caption{
Results on Switchboard 300 hours training combining the representation from different models.
}
\label{tab:swb_300_different}
\end{table}

\begin{table}[t]
\setlength{\aboverulesep}{1pt}
\setlength{\belowrulesep}{1pt}
\setlength\tabcolsep{8pt}
\begin{tabular}{llcc}
\toprule[\heavyrulewidth]
\multicolumn{2}{c}{KD Strategy} & SWB & CH \\
\midrule
\midrule
\multicolumn{2}{l}{No KD (1$^{\textrm{st}}$ itr)} & 10.6 & 6.6 \\
\multicolumn{2}{l}{No KD (2$^{\textrm{nd}}$ itr)} & 10.9 & 6.3 \\
\multicolumn{2}{l}{Uniform 1L} & \textbf{10.1} & 6.5 \\
\multicolumn{2}{l}{Uniform 2L + Context (10\% Mask)} & 10.3 & \textbf{6.2} \\
\bottomrule[\heavyrulewidth]
\end{tabular}
\caption{
Results on Switchboard 2000 hours training.
}
\label{tab:swb_2000_best}
\end{table}

In Table \ref{tab:swb_300_intermediate}, we show the training results on the Switchboard 300 hours corpus distilling intermediate layers of pretrained BERT-Base with various layer mapping strategies.

From these results, we can verify that knowledge distillation has a significant improvement over the no distillation baselines.
We can also confirm that transferring multiple representations leads to better results: specifically, transferring 2 or 3 layers led to the best performance, while using more than 3 layers generally led to worse results. This can be explained by the capacity limitation of the current ASR baseline.

Interestingly, we found that all layer mapping strategies perform almost equivalently well, i.e. even transferring the first few layers or random layers of LLM led to competitive results.
This indicates that linguistic features useful for ASR can be learnt from both lower and higher layers.
In the rest of this paper, we focus on the uniform strategy as the representative strategy and improve upon this baseline.

In Table \ref{tab:swb_300_context}, we show the results of applying context augmentation, i.e. we add the last layer representation of pretrained BERT-Base with additional context as input.
Here, we can verify that context augmentation provides consistent improvement over the uniform strategy with 1 or 2 layers.
In addition, applying random masking to the context tokens yields further improvement: this indicates that transferring diverse, dynamic representations of LLM can be important to prevent overfitting and facilitate student learning.

In Table \ref{tab:swb_300_different}, we show the results of combining the representation from different models. 
From these numbers, we can confirm that a different model, either trained on different dataset (BERT-Finetuned) or with different architecture (DistilBERT and BERT-Large), can provide complementary benefit to the primary model (BERT-Base).
Interestingly, we found that using DistilBERT as the teacher outperforms larger and more powerful BERT-Large.
This can be attributed to the capacity gap between the student and teacher, emphasizing the importance of choosing the appropriate LLM teacher size depending on the ASR student model size.

Finally, we report the results of Switchboard 2000 hours training in Table \ref{tab:swb_2000_best}.
For multiple representation transfer, we chose the best configuration from the 300 hours experiments, namely Uniform 2L + Context (10\% Mask).

While the distillation methods perform decently well, we could only confirm a relatively small improvement over the no distillation baselines, either based on the single or multiple representation transfer.
This may be due to the student already learning sufficient linguistic knowledge in the high-resource setting, and the benefit of LLM knowledge distillation is more salient when the ASR training data is limited.

\section{Conclusion}
\label{sec:conclusion}

In this study, we proposed a novel approach of obtaining and transferring multiple representations of LLM into a transducer ASR system.
From the Switchboard 300 hours experiments, we confirmed the advantage of transferring multiple intermediate representations of LLM based on various layer mapping strategies.
In addition, we examined the effect of combining context augmentation and different model representations, which led to further improvement over the uniform mapping baselines.
Finally, from the Switchboard 2000 hours experiment, we observed the benefit of LLM distillation diminishes for both single and multiple representation transfer, and improving upon such stronger ASR baselines remains as a challenging open problem left as future work.

\vfill\pagebreak
\small


\bibliographystyle{IEEEbib}
\bibliography{strings,refs}

\end{document}